\documentclass[11pt,a4paper]{article}
\usepackage[utf8]{inputenc}
\usepackage[]{graphicx}

\title{Humor in Collective Discourse: Unsupervised Funniness Detection in the New Yorker Cartoon Caption Contest}

\small

\author{Dragomir Radev$^1$, Amanda Stent$^2$, Joel Tetreault$^2$, Aasish Pappu$^2$\\
  Aikaterini Iliakopoulou$^3$, Agustin Chanfreau$^3$, Paloma de Juan$^2$\\ 
  Jordi Vallmitjana$^2$, Alejandro Jaimes$^2$, Rahul Jha$^1$, Bob Mankoff$^4$\\
$^1$ University of Michigan\\
$^2$ Yahoo! Labs\\
$^3$ Columbia University\\
$^4$ The New Yorker\\
\small (radev@umich.edu, stent@yahoo-inc.com, tetreaul@yahoo-inc.com\\
\small aasishkp@yahoo-inc.com, ai2315@columbia.edu, ac3680@columbia.edu\\ 
\small pdejuan@yahoo-inc.com, jvallmi@yahoo-inc.com, ajaimes@yahoo-inc.com\\
\small rahuljha@umich.edu, bob\_mankoff@newyorker.com)}
\date{April 2015}

\begin{document}
\maketitle

\begin{abstract}
The New Yorker publishes a weekly captionless cartoon. More than 5,000 readers submit captions for it. The editors select three of them and ask the readers to pick the funniest one. We describe an experiment that compares a dozen automatic methods for selecting the funniest caption. We show that negative sentiment, human-centeredness, and lexical centrality most strongly match the funniest captions, followed by positive sentiment. These results are useful for understanding humor and also in the design of more engaging conversational agents in text and multimodal (vision+text) systems. As part of this work, a large set of cartoons and captions is being made available to the community.

\end{abstract}

\section{Introduction}

\begin{figure}[h]
\centering
\includegraphics[scale=1]{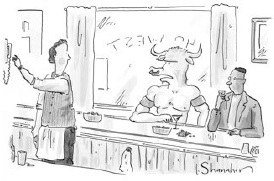}
\caption{Cartoon number 31}
\label{cartoon31}
\end{figure}

\begin{figure}[h]
\centering
\includegraphics[scale=1]{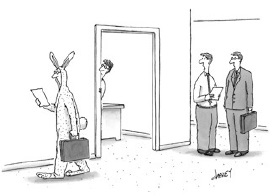}
\caption{Cartoon number 32}
\label{cartoon32}
\end{figure}

The New Yorker Cartoon Caption Contest has been running for more than 10 years. Each week, the editors post a cartoon (cf. Figures~\ref{cartoon31} and~\ref{cartoon32}) and ask readers to come up with a funny caption for it. They pick the top 3 submitted captions and ask the readers to pick the weekly winner. The contest has become a cultural phenomenon and has generated a lot of discussion as to what makes a cartoon funny (at least, to the readers of the New Yorker). 
In this paper, we take a computational approach to studying the contest to gain insights into what differentiates funny captions from the rest. 
We developed a set of unsupervised methods for ranking captions based on features such as originality, centrality, sentiment, concreteness, grammaticality, human-centeredness, etc. We used each of these methods to independently rank all captions from our corpus and selected the top captions for each method. Then, we performed Amazon Mechanical Turk experiments in which we asked Turkers to judge which of the selected captions is funnier.

\section{Related Work}


In early work, Mihalcea and Strapparava \cite{Mihalcea&Strapparava05} investigate whether classification techniques can distinguish between humorous and non-humorous text. Training data consisted of humorous one-liners (15 words or less), and non-humorous one-liners, which are derived from Reuters news titles, proverbs, and sentences from the British National Corpus. They looked at features such as  alliteration, antonymy and adult slang.

Mihalcea and Pullman \cite{conf/cicling/MihalceaP07} took this work further. They looked at four semantic classes relevant to human-centeredness: persons, social groups, social relationships, and personal pronouns. They showed that social relationships and personal pronouns have high prevalence in humor. Mihalcea and Pullman also looked at sentiment; they found that humor tends to have a strong negative orientation (especially in the case of long satirical text, but regular text also shows some tendency toward the negative). Reyes et al. \cite{conf/iicai/ReyesRB09} used these same features as well as others to build a humor taxonomy.

Raz \cite{raz2012automatic} classified tweets by type and topic, while Barberi \cite{barbieriautomatic} focused on classifying tweets into Irony, Education, Humour, and Politics. Zhang et al \cite{zhang2014recognizing}, also looking at tweets, used a set of manually crafted features based on influential humor theories, linguistic norms, and affective dimensions.

  Our work differs from previous research in several ways. First, most previous work has focused on automatically distinguishing between humorous and non-humorous text. In our case, the goal is to rank humorous texts (and assess {\it why} they are funny), not perform binary classification. Second, we're not aware of any work that deals specifically with cartoon captions, and although our methods are not specific to captions, we include features based on the objects depicted in the cartoons.

\section{Data}

We have access to a corpus of more than 2M captions for more than 400 contests run since 2005. For our experiments we picked a subset of 50 cartoons and 298,224 captions. Our data set includes, for each contest, the following:

\begin{itemize}
\item the cartoon itself
\item 5,000+ captions, tokenized using ClearNLP 2.0 \cite{choi.palmer:2012b}
\item the three selected captions, including the winning caption
\item the most frequent n-grams in the captions
\item manually labeled objects that are visible in the cartoon
\item tfidf scores for all captions
\item ``antijokes" from two sites (AlInLa\footnote{http://alinla.blogspot.com/} and Radosh\footnote{http://www.radosh.net/}), devoted to ``unfunny" captions
\end{itemize}

\section{Experimental Setup}

We developed more than a dozen unsupervised methods for ranking the submissions for a given contest. As controls, we use the three captions selected by the editors of the New Yorker as well as antijokes. For all methods, we broke ties randomly. Some of our methods can be used in two different directions (e.g., {\tt CU2} favors the most positive captions whereas {\tt CU2R} the most negative ones). The methods and baselines are split into five groups: OR$=$originality based, GE$=$generic, CU$=$content, NY$=$original New Yorker contest, CO$=$control.

\begin{itemize}
\item ({\tt OR1 \& OR1R}) similarity to contest centroid
\item ({\tt OR2 \& OR2R}) highest/lowest lexrank
\item ({\tt OR3 \& OR3R}) largest/smallest cluster
\item ({\tt OR4}) highest average tfidf

\item ({\tt CU1}) presence of Freebase entities \cite{Bollacker2008}
\item ({\tt CU2 \& CU2R}) caption sentiment
\item ({\tt CU3}) human-centeredness

\item ({\tt GE1}) most syntactically complex 
\item ({\tt GE2}) most concrete (i.e., refers to objects present in the cartoon)
\item ({\tt GE3 \& GE3R}) unusually formatted text

\item ({\tt NY1}) first place official
\item ({\tt NY2}) second place official
\item ({\tt NY3}) third place official

\item ({\tt CO2}) antijokes
\end{itemize}

\subsection{Originality-based methods}

We built a lexical network out of the captions for each contest.  We used LexRank \cite{Erkan&Radev04} 
to identify the most central caption in each contest (method {\tt OR1}) and the one with the highest lexrank score (method {\tt OR2}). We also used a graph clustering method \cite{Blondel&al.08}, previously used in King et al. \cite{King&al.13a}, to cluster the captions in each contest thematically; the sizes of these clusters comprise  method {\tt OR3}. The tfidf scores used to build the lexical network are used in method {\tt OR4}.

\begin{figure*}[htp]
\scriptsize
\begin{verbatim}
 0 0    if that 's theseus , i 'm not here .
 1 0    if it 's theseus , tell him i 'll be back in the labyrinth just as soon as happy hour is over .
 2 0    if that 's theseus , i just left .
 3 0    if it 's theseus , tell him to get lost .
 4 1    if that 's elsie , you have n't seen me .
 5 2    if that 's bessie , tell her i 've moooooved on !
 6 3    if its my wife , tell her i 'm in a china shop .
 7 3    i got kicked out of the china shop .
 8 5    if that 's merrill lynch , tell them i quit and went to pamplona .
 9 5    if that 's my wife , tell her i went to pamplona .
10 4    if it 's my wife , tell her that i ran into an old minotaur friend .
11 4    if that 's my wife tell her i 'll be home in a minotaur .
12 4    jeez ! what 's a minotaur got to do to get a drink around here ? 
13 4    if i hear that ' a guy and a minotaur go into a bar ' joke one more time ... 
14 5    if that 's merrill lynch , tell them i 'll be back when i 'm good and ready . 
15 5    if it 's my wife , i was working late on a merrill-lynch commercial . 
16 5    if that 's my cow , tell her i left for pamplona . 
17 3    this 'll be the last one . i need to get back to the china shop . 
18 6    if that 's my matador , tell him i 'm not here . 
19 5    if that 's merrill or lynch , tell ' em i 'm not here . 
\end{verbatim}
\caption{Subset of the captions for contest number 31, labeled by thematical cluster (column 2). 0 - theseus, 1 - elsie, 2 - bessie, 3 - china shop, 4 - minotaur, 5 - merrill lynch, 6 - matador.}
\label{mini-corpus}
\end{figure*}

\begin{figure}[htpb]
\centering
\includegraphics[scale=.5]{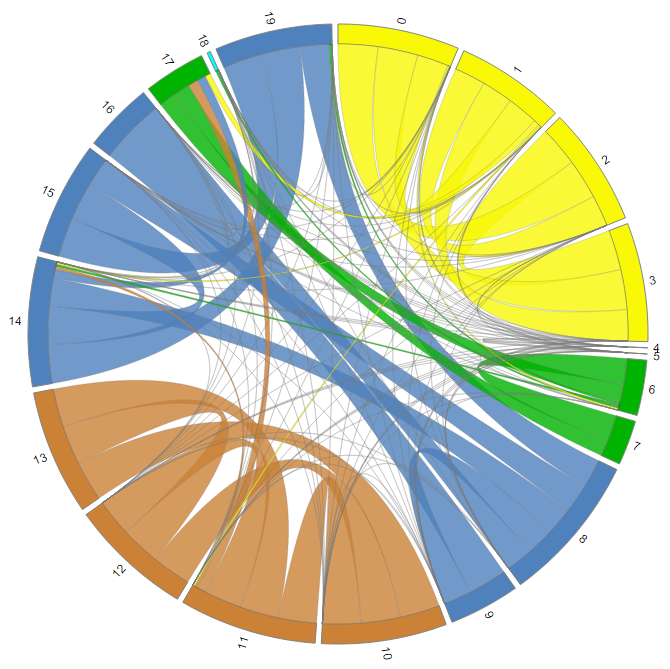}
\caption{Clustering of the mini corpus}
\label{circularplot}
\end{figure}

Figure~\ref{circularplot} shows the pairwise similarities for the
captions in the mini-corpus.  The seven clusters are identified by the
Louvain method. Solid lines represent high cosine similarity between a
pair of captions.

\begin{figure}[htpb]
\centering
\includegraphics[scale=.5]{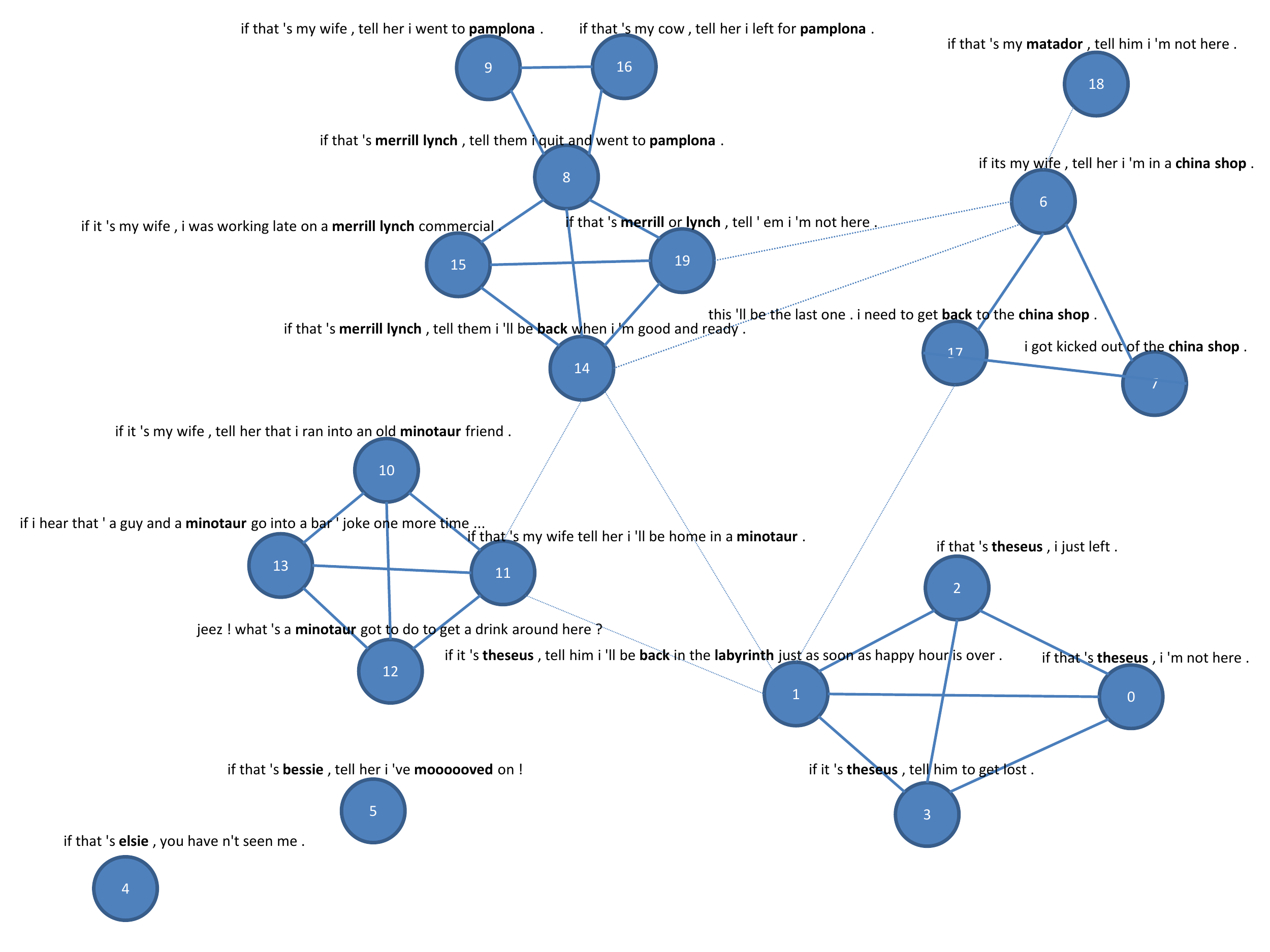}
\caption{Lexical network for contest 31.}
\label{minotaurgraph}
\end{figure}

The captions in the mini-corpus are shown in
Figure~\ref{mini-corpus}. The seven clusters in
Figure~\ref{minotaurgraph} are identified by the Louvain method. Solid
lines represent high cosine similarity between a pair of captions.

\subsection{Content-based methods}

For {\tt CU1}, we annotated the captions for Freebase entities by querying noun-phrases (within a caption) over Freebase indexed entities. We scored each caption using idf $*$ Freebase score, where the Freebase score captures relevance.

To compute the sentiment polarity of each caption (method {\tt CU2}), we used Stanford CoreNLP \cite{manning-EtAl:2014:P14-5} to annotate each sentence with its sentiment from 0 (very negative) to 4 (very positive). Only 13.20\% had positive polarity; 51.09\% had negative polarity, and the rest were neutral. 

For human-centeredness (method {\tt CU3}), we followed the method described in Mihalcea and Pullman \cite{conf/cicling/MihalceaP07}. 
We used WordNet \cite{Miller:1995:WLD:219717.219748} to list all the word forms derived from the \{\textit{person, individual, someone, somebody, mortal, human, soul}\} synset (``people'' set), as well as those belonging to the \{\textit{relative, relation}\} synset (``relatives'' set). We excluded personal pronouns, as 75.96\% of the captions contained at least one. We also accounted for any proper names as part of the ``people'' set. 25.33\% of the captions mentioned at least one ``person'', but only 3.60\% contained a word from the ``relatives'' set.

\subsection{Generic methods}

We computed syntactic complexity ({\tt GE1}) using \cite{charniak-johnson:2005:ACL}. For concreteness ({\tt GE2}), two of the authors of this paper labeled all the objects in each of the 50 cartoons used in our evaluation. We then computed how often any of those objects were referred to (with a nominal NP) in each caption. We computed {\tt GE3} by counting punctuation marks and unusually formatted (e.g. very long) words in each caption.

\begin{table*}[htpb]
\centering
\scriptsize
\begin{tabular}{|l|c|r||r|r||r|r||r|r|r|}
\hline
{\bf Category} & {\bf Code} & {\bf Method} & {\bf $n_4$} & {\bf $s_4$} & {\bf $n_3$} & {\bf $s_3$} & {\bf $n$} & {\bf $s$} \\
\hline
Centrality & OR1R & least similar to centroid & 308 & -2.73 & 453 & -2.14 & 846 & -1.26\\
& {\bf OR2} & highest lexrank & 302 & {\bf 1.39} & 457 & {\bf 1.11} & 846 & {\bf 0.59} \\
& OR2R & smallest lexrank & 317 & -0.61 & 450 & -0.58 & 846 & -0.29 \\
& OR3R & small cluster & 468 & -4.40 & 581 & -3.94 & 848 & -2.85 \\
& OR4 & tfidf & 474 & -4.93 & 596 & -4.36 & 850 & -3.24\\
\hline
New Yorker & {\bf NY1} & official winner & 314 & {\bf 3.57} & 466 & {\bf 2.96} & 847 & {\bf 1.78}\\
& {\bf NY2} & official runner up & 330 & {\bf 3.24} & 463 & {\bf 2.60} & 845 & {\bf 1.54}\\
& {\bf NY3} & official third place & 276 & {\bf 2.29} & 435 & {\bf 1.57} & 842 & {\bf 0.89}\\
\hline
General & GE1 & syntactically complex & 268 & -0.10 & 406 & -0.14 & 846 & -0.70\\
& GE2 & concrete & 259 & -0.33 & 427 & -0.41 & 844 & -0.26\\
& GE3R & well formatted & 296 & 0.81 & 446 & 0.61 & 846 & 0.31\\
\hline
Content & CU1 & freebase & 290 & 0.26 & 424 & 0.17 & 840 & 0.07\\
& {\bf CU2} & positive sentiment & 268 & {\bf 1.21} & 396 & {\bf 0.83} & 836 & {\bf 0.46} \\
& {\bf CU2R} & negative sentiment & 298 & {\bf 1.69} & 445 & {\bf 1.30} & 826 & {\bf 0.70} \\
& {\bf CU3} & people & 276 & {\bf 1.45} & 409 & {\bf 1.24} & 834 & {\bf 0.68} \\
\hline
Control & CO2 & antijoke & 259 & 0.27 & 394 & -0.04 & 822 & -0.09\\
\hline
\end{tabular}
\caption{Comparison between the methods. Score $s_4$ corresponds to pairs for which the seven judges agreed more significantly (a difference of 4+ votes). Score $s_3$ requires a difference of 3+ votes. Score $s$ includes all pairs (about 850 per method, minus a small number of errors).
The best methods (CU2R, CU3, OR2, and CU2)  are in bold. }
\label{results}
\end{table*}

\section{Evaluation}

We used Amazon Mechanical Turk (AMT) to compare the outputs of the different methods and the baselines. Each AMT HIT consisted of one cartoon as well as two captions, A and B (produced by one of the 18 methods and baselines). The turkers had to determine which of the two captions is funnier. They were given four options - ``A is funnier'', ``B is funnier'', ``both are funny'', ``neither is funny''. They did not know which method was used to produce caption A or B. All pairs of captions from our methods were compared for each cartoon, and each HIT (pair) was assessed by 7 Turkers.

We report on three evaluations in Table~\ref{results}. Each evaluation ({$n_{i}$, $s_{i}$} pair) corresponds to the number of votes in favor of the given method minus the number of votes against.  So the first set corresponds to pairs in which, out of seven judges, there was a difference of at least 4 votes in favor of one or the other caption. This level of significant agreement happened in 5,594/15,154 cases (36.9\% of the time). A difference of at least 3 votes happened in 8,131/15,154 pairs (53.6\%). The third evaluation corresponds to all pairwise comparisons, including ties.  $n_{i}$ refers to the number of times the above 
constraint for $i$ is met and score $s_{i}$ is calculated by averaging the number of votes in favor minus the number of votes against for each $n_{i}$.
The probability that a random process will generate a difference of at least 4 votes (excluding ties) is 12.5\%.


\section{Conclusion}

We compared over a dozen methods for selecting the funniest caption among 5,000 submissions to the New Yorker caption contest. Using side by side funniness assessments from AMT, we found that the methods that consistently select funnier captions are negative sentiment, human-centeredness, and lexical centrality. Not surprisingly, knowing the traditions of the New Yorker cartoons, negative captions were funnier than positive captions. Captions that relate to people were consistently deemed funnier. The first two methods (negative sentiment and human-centeredness) are consistent with the findings in Mihalcea and Pullman \cite{conf/cicling/MihalceaP07}. More interestingly, we also showed that captions that reflect the collective wisdom of the contest participants outperformed semantic outliers. The next two strongest features were positive sentiment and proper formatting.

We are making our corpus public for research and for a shared task on funniness detection. The corpus includes our 50 selected cartoons, more than 5,000 captions per cartoon, manual annotations of the entities in the cartoons, automatically extracted topics from each contest, and the funniness scores. 

\section{Future Work}

In this paper, we used unsupervised methods for funniness detection. We will next explore supervised and ensemble methods. (However, ensemble methods may not work for this task as captions may be funny in different ways; for example, of two equally funny captions, one may be funny-absurd and the other funny-ironic.) We will also explore pun recognition (e.g., "Tell my wife I'll be home in a {\em minotaur}."), other creative uses of language, as well as more semantic features. 




\begin{thebibliography}{10}

\bibitem{barbieriautomatic}
Francesco Barbieri and Horacio Saggion.
\newblock Automatic detection of irony and humour in twitter.
\newblock In {\em Proceedings of the International Conference on Computational
  Creativity}, 2014.

\bibitem{Blondel&al.08}
Vincent~D Blondel, Jean-Loup Guillaume, Renaud Lambiotte, and Etienne Lefebvre.
\newblock Fast unfolding of communities in large networks.
\newblock {\em Journal of Statistical Mechanics: Theory and Experiment},
  2008(10), 2008.

\bibitem{Bollacker2008}
Kurt Bollacker, Colin Evans, Praveen Paritosh, Tim Sturge, and Jamie Taylor.
\newblock Freebase: a collaboratively created graph database for structuring
  human knowledge, 2008.

\bibitem{charniak-johnson:2005:ACL}
Eugene Charniak and Mark Johnson.
\newblock Coarse-to-fine n-best parsing and maxent discriminative reranking.
\newblock In {\em Proceedings of the ACL}, 2005.

\bibitem{choi.palmer:2012b}
Jinho~D. Choi and Martha Palmer.
\newblock Fast and robust part-of-speech tagging using dynamic model selection.
\newblock In {\em Proceedings of the ACL}, 2012.

\bibitem{Erkan&Radev04}
G\"{u}ne\c{s} Erkan and Dragomir~R. Radev.
\newblock Lexrank: Graph-based centrality as salience in text summarization.
\newblock {\em Journal of Artificial Intelligence Research}, 22:457--479, 2004.

\bibitem{King&al.13a}
Benjamin King, Rahul Jha, Dragomir~R. Radev, and Robert Mankoff.
\newblock Random walk factoid annotation for collective discourse.
\newblock In {\em Proceedings of The ACL}, 2013.

\bibitem{manning-EtAl:2014:P14-5}
Christopher~D. Manning, Mihai Surdeanu, John Bauer, Jenny Finkel, Steven~J.
  Bethard, and David McClosky.
\newblock The {Stanford} {CoreNLP} natural language processing toolkit.
\newblock In {\em Proceedings of the ACL}, pages 55--60, 2014.

\bibitem{conf/cicling/MihalceaP07}
Rada Mihalcea and Stephen~G. Pulman.
\newblock Characterizing humour: An exploration of features in humorous texts.
\newblock In {\em Proceedings of CICLing}, 2007.

\bibitem{Mihalcea&Strapparava05}
Rada Mihalcea and Carlo Strapparava.
\newblock Making computers laugh: Investigations in automatic humor
  recognition.
\newblock In {\em Proceedings of HLT/EMNLP}, 2005.

\bibitem{Miller:1995:WLD:219717.219748}
George~A. Miller.
\newblock {WordNet}: A lexical database for {English}.
\newblock {\em Communications of the ACM}, 38(11):39--41, Nov. 1995.

\bibitem{raz2012automatic}
Yishay Raz.
\newblock Automatic humor classification on {Twitter}.
\newblock In {\em Proceedings of NAACL/HLT}, 2012.

\bibitem{conf/iicai/ReyesRB09}
Antonio Reyes, Paolo Rosso, and Davide Buscaldi.
\newblock Evaluating humorous features: Towards a humour taxonomy.
\newblock In {\em Proceedings of the Indian International Conference on
  Artificial Intelligence}, 2009.

\bibitem{zhang2014recognizing}
Renxian Zhang and Naishi Liu.
\newblock Recognizing humor on twitter.
\newblock In {\em Proceedings of the ACM International Conference on
  Information and Knowledge Management}, 2014.

\end{thebibliography}

\end{document}